\documentclass[letterpaper, 10 pt, conference]{ieeeconf}  

\IEEEoverridecommandlockouts                              

\title{\LARGE \bf
Dropping the D: RGB-D SLAM Without the Depth Sensor
}

\author{Mert Kiray$^{1, 2, 3*}$, Alican Karaomer$^{1*}$, and Benjamin Busam$^{1,2,3}$ 
\thanks{$*$equal contribution}%
\thanks{$^{1}$Mert Kiray, Alican Karaomer, Benjamin Busam are with Technical University of Munich, Munich, Germany {\tt\small mert.kiray@tum.de}}%
\thanks{$^{2}$Mert Kiray and Benjamin Busam are with 3Dwe.ai, Munich, Germany}%
\thanks{$^{3}$Mert Kiray and Benjamin Busam are with Munich Center for Machine Learning}%
\thanks{$^{4}$This work has been submitted to the IEEE for possible publication. Copyright may be transferred without notice, after which this version may no longer be accessible.}
}

\usepackage{hyperref}
\usepackage{graphicx}
\usepackage{amsmath} 
\usepackage[bb=dsserif]{mathalpha}
\usepackage{xspace}
\usepackage{booktabs}

\usepackage{pgfplots}
\usepackage{pgfplotstable}
\usepackage{tabularx}
\pgfplotsset{compat=1.18} 

\usepackage{float}   
\usepackage{multirow}              
\usepackage[table,xcdraw]{xcolor}  
\usepackage{colortbl}              

\let\labelindent\relax
\usepackage{enumitem}   

\usepackage{dblfloatfix}
\usepackage{adjustbox}
\usepackage{subcaption}
\usepackage{booktabs,siunitx}
\usepackage{listings}
\usepackage{color}

\definecolor{codegreen}{rgb}{0,0.6,0}
\definecolor{codegray}{rgb}{0.5,0.5,0.5}
\definecolor{codepurple}{rgb}{0.58,0,0.82}
\definecolor{backcolour}{rgb}{0.95,0.95,0.92}

\def\approach{DropD-SLAM\xspace}

\begin{document}

\makeatletter
\g@addto@macro\@maketitle{
\vspace{-10pt}
\begin{figure}[H]
\setlength{\linewidth}{\textwidth}
\setlength{\hsize}{\textwidth}
\centering
\includegraphics[width=\linewidth]{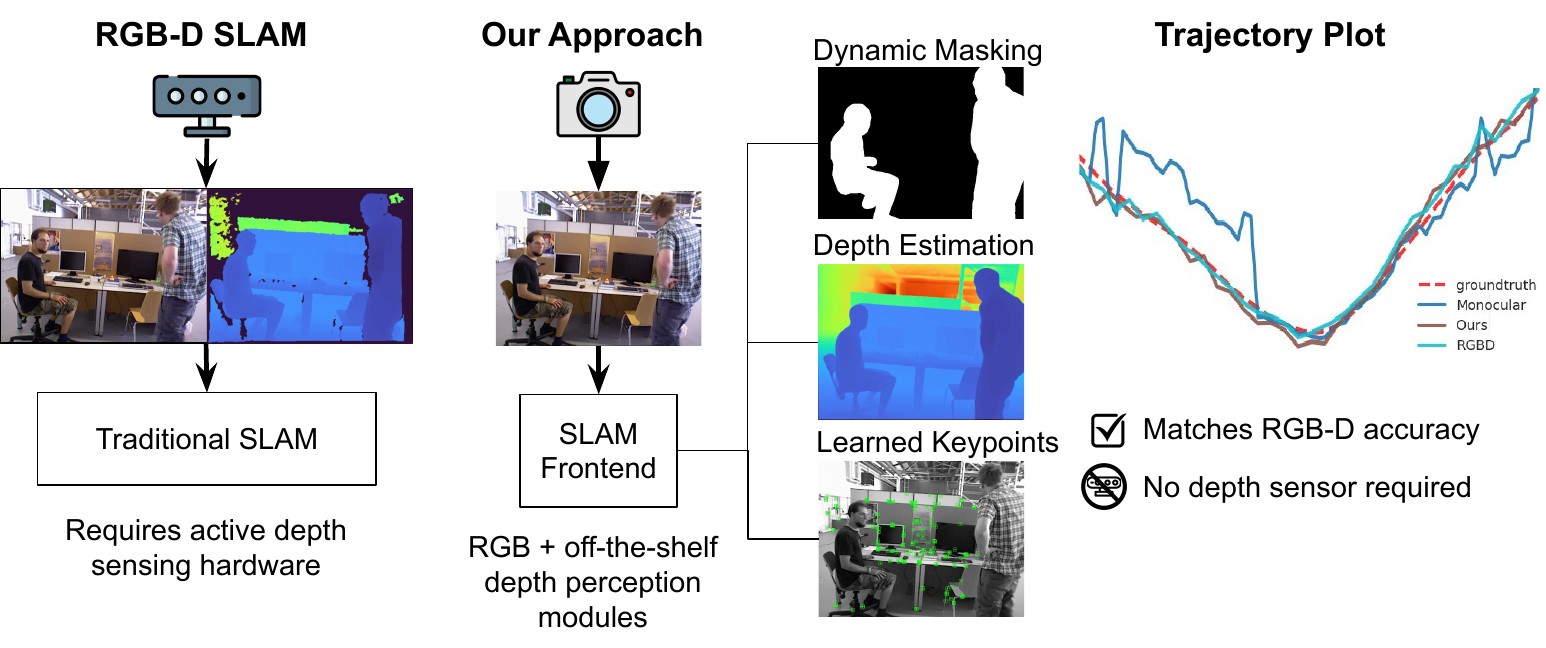}
\vspace{-10pt}
\caption{Conceptual comparison between traditional RGB-D SLAM and our proposed \approach. 
Conventional pipelines require active depth sensing for scale and robustness, 
while \approach achieves comparable performance from a single RGB input by leveraging pretrained modules for depth, features, and instance segmentation.}
\label{fig:teaser}
\end{figure}
}
\makeatother
\maketitle
\setcounter{figure}{1}
\thispagestyle{empty}
\pagestyle{empty}
\begin{abstract}
    We present \approach, a real-time monocular SLAM system that achieves RGB-D-level accuracy without relying on depth sensors. The system replaces active depth input with three pretrained vision modules: a monocular metric depth estimator, a learned keypoint detector, and an instance segmentation network. Dynamic objects are suppressed using dilated instance masks, while static keypoints are assigned predicted depth values and backprojected into 3D to form metrically scaled features. These are processed by an unmodified RGB-D SLAM back end for tracking and mapping. On the TUM RGB-D benchmark, \approach attains 7.4\,cm mean ATE on static sequences and 1.8\,cm on dynamic sequences, matching or surpassing state-of-the-art RGB-D methods while operating at 22\,FPS on a single GPU. These results suggest that modern pretrained vision models can replace active depth sensors as reliable, real-time sources of metric scale, marking a step toward simpler and more cost-effective SLAM systems. Code available at \href{https://tum-pf.github.io/dropd-slam/}{tum-pf.github.io/dropd-slam}
    \end{abstract}
\section{Introduction}

Monocular SLAM remains attractive for its simplicity and minimal hardware requirements, yet it continues to face two persistent limitations: scale ambiguity and sensitivity to dynamic environments. Without metric depth, monocular pipelines suffer from scale drift and unstable landmark initialization, especially during early tracking. In addition, dynamic objects violate the static-world assumption underlying most SLAM formulations, producing incorrect correspondences and degraded pose estimates. These challenges have traditionally motivated the use of active depth sensors to achieve robust, metric-scale SLAM in real-world applications.

Active sensing modalities such as RGB-D cameras and LiDAR provide metrically scaled depth and improved robustness to scene dynamics. However, they introduce cost, power consumption, and system complexity, and they are susceptible to degradation in outdoor settings, on reflective materials, or in low-texture regions~\cite{jung2024housecat6d}.

Recent advances in vision models present a promising alternative. Pretrained monocular depth estimators~\cite{yang2024depth,piccinelli2025unidepthv2} now deliver dense metric predictions with strong generalization across scenes and intrinsics. Learned keypoint detectors such as Key.Net~\cite{barroso2019key} improve repeatability under motion blur and low texture, while instance segmentation networks such as YOLOv11~\cite{khanam2024yolov11} provide efficient localization of dynamic objects. Together, these modules create an opportunity to recover metric structure and suppress dynamic content using only monocular RGB input.

We propose \approach, a real-time monocular SLAM system that achieves RGB-D-level accuracy without relying on depth sensors. The method replaces the active depth stream of a classical SLAM front end with three pretrained modules: a monocular metric depth estimator, a learned keypoint detector, and an instance segmentation model. Dynamic regions are identified with instance-level masks and filtered using morphological dilation, while the remaining keypoints are assigned predicted metric depth and backprojected into 3D to initialize map points at absolute scale. The resulting features are processed by a standard RGB-D SLAM back end, which remains unmodified (see~\autoref{fig:teaser}). All components run in real time without task-specific training or fine-tuning, enabling zero-shot deployment across diverse environments.

Unlike prior learned SLAM approaches that rely on end-to-end optimization, scene-specific adaptation, or custom back ends, \approach preserves the classical geometric pipeline. By treating pretrained models as modular perception units, it achieves strong performance in both static and dynamic indoor settings while maintaining architectural simplicity and compatibility with existing systems.

\textbf{Contributions.}
\begin{itemize}[leftmargin=*]
\item We propose \approach, a real-time monocular SLAM system that integrates pretrained monocular depth~\cite{yang2024depth,piccinelli2025unidepthv2}, learned keypoints~\cite{barroso2019key}, and instance segmentation~\cite{khanam2024yolov11} into a modular front end that remains fully compatible with standard RGB-D back ends~\cite{mur2017orb}, which allows metric scale without the need for depth sensors.
\item We introduce a dynamic object filtering strategy based on instance-level segmentation with morphological dilation and we show that learned keypoints improve robustness under motion blur and texture-poor conditions.
\item Through ablation studies we identify temporal depth consistency rather than per-frame accuracy as the dominant factor for monocular SLAM performance, which provides a new perspective on the role of pretrained depth models.
\item We achieve state-of-the-art results on the TUM RGB-D benchmark~\cite{sturm2012benchmark} with 7.4\,cm mean ATE on static sequences and 1.8\,cm on dynamic sequences, and our system runs in real time (22\,FPS) on a single GPU.
\end{itemize}

\begin{figure*}[t]
    \centering
    \includegraphics[width=\linewidth]{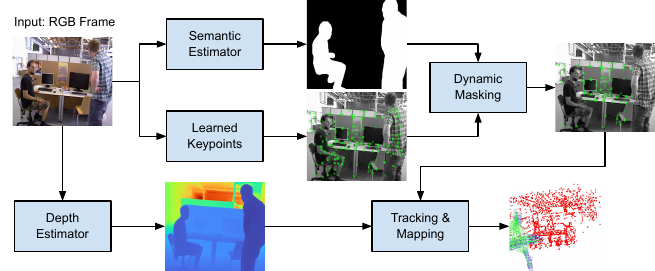}
\caption{Overview of \approach. Each RGB frame is processed in parallel by three pretrained modules: 
(i) metric depth estimation, (ii) instance segmentation, and (iii) keypoint detection. 
Instance masks filter dynamic objects, and static keypoints are backprojected with predicted depth to form metrically scaled 3D features. 
These are passed to an unmodified RGB-D SLAM back end for tracking, mapping, and loop closure in real time.}
    \label{fig:pipeline}
\end{figure*}

\section{Related Work}

\subsection{Classical and Learned SLAM Systems}
Classical SLAM approaches are broadly categorized into feature-based and direct methods. Feature-based pipelines, exemplified by the ORB-SLAM series~\cite{mur2015orb, mur2017orb, campos2021orb}, rely on sparse keypoint detection and matching. Direct and semi-direct methods such as LSD-SLAM~\cite{engel2014lsd}, DSO~\cite{engel2017direct}, and SVO~\cite{forster2014svo} minimize photometric error over pixel intensities, which improves performance in low-texture scenes. To enhance feature quality, learned detectors such as SuperPoint~\cite{detone2018superpoint} and Key.Net~\cite{barroso2019key} have been introduced to increase repeatability under challenging conditions. However, these approaches typically require stereo or active depth sensing to recover metric scale and often assume static environments.

Recent learned SLAM pipelines incorporate data-driven priors through different mechanisms. Some methods replace components of the classical pipeline. For example, DROID-SLAM~\cite{teed2021droid} uses recurrent networks for correspondence refinement, while other approaches optimize compact scene representations such as depth codes~\cite{bloesch2018codeslam, czarnowski2020deepfactors}. More recent work explores representations like Gaussian splatting~\cite{matsuki2024gaussian, sandstrom2025splat} and integrates learned cues for dynamic content~\cite{zheng2025wildgs, zhang2024monst3r}. These methods often improve accuracy but rely on custom back ends or offline training.

\subsection{Monocular Depth Estimation}
Early supervised and self-supervised monocular depth estimators, including MonoDepth~\cite{monodepth17}, MiDaS~\cite{ranftl2020towards}, and ZoeDepth~\cite{bhat2023zoedepth}, provide strong relative depth but lack metric scale. More recent models target zero-shot metric estimation with improved generalization. DepthAnythingV2~\cite{yang2024depth} and UniDepthV2~\cite{piccinelli2025unidepthv2} produce fast, dense predictions by leveraging large-scale training and explicit intrinsics encoding. Other efforts emphasize cross-camera consistency~\cite{hu2024metric3d}, high-resolution output~\cite{bochkovskii2024depth}, or temporal stability~\cite{chen2025video}. While these models deliver high-quality predictions, their integration into SLAM pipelines remains challenging due to geometric and consistency constraints.

\subsection{SLAM in Dynamic Environments}
Dynamic content disrupts standard SLAM assumptions and creates difficulties for reliable pose estimation. Prior work addresses this challenge using semantic filtering to exclude dynamic object classes~\cite{yu2018ds, bescos2018dynaslam}, geometric or motion-based outlier rejection~\cite{yu2022d}, or learned dynamic models~\cite{jiang2024rodyn, xu2024dg}. Many of these approaches assume access to depth or stereo input and often introduce significant computational overhead.

\subsection{Monocular Depth in SLAM Pipelines}
Several systems have investigated the use of predicted depth in place of active sensors. CNN-SLAM~\cite{tateno2017cnn} demonstrated early integration of single-frame depth predictions for static scenes. Later methods optimized joint representations of depth and pose~\cite{bloesch2018codeslam, czarnowski2020deepfactors}, but typically relied on custom optimization and failed to achieve real-time performance.

Most prior work improves individual aspects of monocular SLAM by incorporating learned depth, more robust feature extraction, or dynamic masking. In contrast, our approach combines all three components into a unified, real-time pipeline that retains a standard geometric back end and enables robust, metrically scaled monocular SLAM in dynamic environments.

\section{Method}
\label{sec:method}

We propose a monocular SLAM front end designed to recover metric scale and remain robust in dynamic environments. The core idea is to transform each incoming RGB image into a set of static, metrically scaled 3D features that can be processed directly by a standard RGB-D SLAM back end. This transformation is implemented via a multi-stage pipeline in which pretrained vision modules extract complementary cues from the input image. These outputs are filtered and fused into a consistent 3D representation suitable for reliable localization and mapping, as illustrated in~\autoref{fig:pipeline}.

\subsection{Front-End Processing Pipeline}
For each RGB image $I_t \in \mathbb{R}^{H \times W \times 3}$ at time $t$, and given the camera intrinsics matrix $K \in \mathbb{R}^{3 \times 3}$, the front end executes a sequence of operations: learned keypoint detection, instance-level segmentation, and monocular metric depth estimation. The goal is to produce a set of high-quality static features with associated depth values, minimizing the influence of independently moving objects and preserving compatibility with a geometric SLAM back end.

Keypoints are detected using a network such as Key.Net~\cite{barroso2019key}, applied to a five-level image pyramid. The detector returns $N$ repeatable 2D locations $\mathbf{u}_i \in \mathbb{R}^2$ that are robust under motion blur and low-texture conditions. Each keypoint is paired with a 256-bit ORB descriptor~\cite{rublee2011orb}, yielding a set of binary descriptors $\mathbf{f}_i \in \{0,1\}^{256}$ that remain compatible with classical SLAM systems. The resulting keypoint–descriptor pairs define the initial feature set:
\begin{equation}
    \label{eq:feature-set}
    \mathcal{F}_t = \{(\mathbf{u}_i, \mathbf{f}_i)\}_{i=1}^N.
\end{equation}

Simultaneously, a pretrained instance segmentation network such as YOLOv11~\cite{khanam2024yolov11} predicts a set of binary masks $\{\mathcal{M}_t(c)\}$, one per detected object instance with class label $c$. These masks identify potentially dynamic regions at pixel resolution. In parallel, a monocular depth estimator such as DepthAnythingV2~\cite{yang2024depth} produces a dense depth map $D_t \in \mathbb{R}^{H \times W}$ in metric units, covering the full image domain.

\subsubsection{Dynamic Feature Filtering}
Moving objects violate the static-scene assumption of SLAM systems and lead to unreliable feature correspondences. To suppress such effects, we define a set of dynamic classes $\mathcal{C}_{\mathrm{dyn}}$ (e.g., \emph{person}, \emph{car}, \emph{bus}) and combine the corresponding instance masks into a binary dynamic mask:
\begin{equation}
    \label{eq:dynamic-mask}
    \mathcal{D}_t(\mathbf{u}) = \mathbb{1}\left[\mathbf{u} \in \bigcup_{c \in \mathcal{C}_{\mathrm{dyn}}} \mathcal{M}_t(c)\right].
\end{equation}
To reduce sensitivity to mask boundaries, $\mathcal{D}_t$ is dilated with a circular structuring element to obtain $\mathcal{D}_t^{\mathrm{dil}}$. All keypoints falling within this dilated region are excluded. The remaining static feature set is defined as:
\begin{equation}
    \label{eq:static-set}
    \mathcal{F}_t^{\mathrm{static}} = \{(\mathbf{u}_i, \mathbf{f}_i) \in \mathcal{F}_t \mid \mathcal{D}_t^{\mathrm{dil}}(\mathbf{u}_i)=0\}.
\end{equation}
This filtering step improves robustness by discarding features associated with dynamic content or uncertain boundaries.

\begin{table*}[t]
  \centering
  \caption{ATE RMSE (m) on the TUM Static subset (\texttt{freiburg1} sequences). Lower is better. X denotes tracking failure.}
  \label{tab:tum_static_full}
  \resizebox{\textwidth}{!}{%
  \begin{tabular}{lcccccccccc}
    \toprule
      & 360 & desk & desk2 & floor & plant & room & rpy & teddy & xyz & Avg. \\
    \midrule
    \multicolumn{11}{l}{\textbf{Calibrated}}\\
    ORB-SLAM3~\cite{campos2021orb}
      & X & \cellcolor{green!20}0.017 & 0.210 & X & 0.034 & X & X & X & \cellcolor{green!20}\textbf{0.009} & -- \\
    DeepV2D~\cite{teed2018deepv2d}
      & 0.243 & 0.166 & 0.379 & 1.653 & 0.203 & 0.246 & 0.105 & 0.316 & 0.064 & 0.375 \\
    DeepFactors~\cite{czarnowski2020deepfactors}
      & 0.159 & 0.170 & 0.253 & 0.169 & 0.305 & 0.364 & 0.043 & 0.601 & 0.035 & 0.233 \\
    DPV-SLAM~\cite{lipson2024deep}
      & 0.112 & \cellcolor{yellow!20}0.018 & 0.029 & 0.057 & \cellcolor{yellow!20}0.021 & 0.330 & 0.030 & 0.084 & \cellcolor{green!20}0.010 & 0.076 \\
    DPV-SLAM++~\cite{lipson2024deep}
      & 0.132 & \cellcolor{yellow!20}0.018 & 0.029 & \cellcolor{yellow!20}0.050 & 0.022 & 0.096 & 0.032 & 0.098 & \cellcolor{green!20}0.010 & 0.054 \\
    GO-SLAM~\cite{zhang2023go}
      & \cellcolor{green!20}0.089 & \cellcolor{green!20}\textbf{0.016} & \cellcolor{yellow!20}0.028 & \cellcolor{green!20}0.025 & 0.026 & \cellcolor{green!20}0.052 & \cellcolor{green!20}\textbf{0.019} & \cellcolor{green!20}0.048 & \cellcolor{green!20}0.010 & \cellcolor{green!20}0.035 \\
    DROID-SLAM~\cite{teed2021droid}
      & 0.111 & \cellcolor{yellow!20}0.018 & 0.042 & \cellcolor{green!20}\textbf{0.021} & \cellcolor{green!20}\textbf{0.016} & \cellcolor{green!20}\textbf{0.049} & \cellcolor{yellow!20}0.026 & \cellcolor{green!20}0.048 & 0.012 & \cellcolor{yellow!20}0.038 \\
    MASt3R-SLAM~\cite{murai2025mast3r}
      & \cellcolor{green!20}\textbf{0.049} & \cellcolor{green!20}\textbf{0.016} & \cellcolor{green!20}\textbf{0.024} & \cellcolor{green!20}0.025 & \cellcolor{green!20}0.020 & 0.061 & 0.027 & \cellcolor{green!20}\textbf{0.041} & \cellcolor{green!20}\textbf{0.009} & \cellcolor{green!20}\textbf{0.030} \\
    \rowcolor{gray!12}\textbf{Ours (RGB-D)}
      & 0.120 & 0.016 & 0.027 & 0.281 & 0.012 & 0.064 & 0.030 & 0.054 & 0.011 & 0.068 \\
    \rowcolor{gray!12}\textbf{Ours (Mono w/DepthAnythingV2~\cite{yang2024depth})}
      & \cellcolor{yellow!20}0.094 & 0.027 & \cellcolor{green!20}0.026 & 0.283 & 0.070 & 0.074 & \cellcolor{green!20}0.023 & 0.150 & 0.012 & 0.083 \\
    \rowcolor{gray!12}\textbf{Ours (Mono w/UniDepthV2~\cite{piccinelli2025unidepthv2})}
      & 0.115 & 0.043 & 0.036 & 0.276 & 0.035 & \cellcolor{yellow!20}0.058 & 0.043 & \cellcolor{yellow!20}0.055 & 0.012 & 0.074 \\
    \bottomrule
  \end{tabular}}
\end{table*}

\subsubsection{Depth Association and 3D Backprojection}
For each $(\mathbf{u}_i, \mathbf{f}_i) \in \mathcal{F}_t^{\mathrm{static}}$, the depth value $d_i = D_t(\mathbf{u}_i)$ is retrieved from the predicted depth map. To reject unreliable predictions, depths are clipped to the interval $[d_{\min}, d_{\max}]$. Each valid keypoint is then backprojected into the 3D camera frame using the pinhole model:
\begin{equation}
    \label{eq:backproj}
    \mathbf{P}_i = d_i K^{-1} [\mathbf{u}_i^\top \ 1]^\top.
\end{equation}
The front end thus produces a set of metrically scaled 3D point–descriptor pairs:
\begin{equation}
    \mathcal{P}_t = \{(\mathbf{P}_i, \mathbf{f}_i)\},
\end{equation}
which serves as input to the SLAM back end. This sparse representation retains compatibility with geometric SLAM while embedding absolute scale from the depth network.

\subsection{Back End Tracking and Mapping}
The set $\mathcal{P}_t$ is passed to the RGB-D interface of an unmodified SLAM system. We adopt ORB-SLAM3~\cite{campos2021orb}, which performs tracking, mapping, and loop closure within a standard geometric optimization framework.

Camera tracking is formulated as a Perspective-n-Point (PnP) problem within a RANSAC loop, aligning the 3D points in $\mathcal{P}_t$ with their 2D projections to estimate the current pose. Mapping proceeds by instantiating new 3D map points when parallax and visibility conditions are met. Loop closures are identified via bag-of-words retrieval over ORB descriptors and verified geometrically; verified matches trigger global pose graph optimization and bundle adjustment to refine the trajectory and structure.

A key design choice is the treatment of predicted depth during optimization. Depth values are treated as fixed priors rather than as variables to be optimized. Each depth introduces a unary residual with variance $\sigma_d^2$ in bundle adjustment, encoding the assumed confidence of the depth estimator. This formulation stabilizes global scale while allowing the optimizer to refine structure via multi-view constraints—without requiring any modification to the back end.

\section{Experiments}
\label{sec:experiments}

We evaluate our system on the TUM RGB-D benchmark~\cite{sturm2012benchmark}, using both static and dynamic indoor sequences. Experiments assess overall SLAM accuracy and isolate the contributions of monocular depth estimation, dynamic object filtering, and learned keypoint detection. Unless otherwise noted, results are reported as the root-mean-square error (RMSE) of Absolute Trajectory Error (ATE) in meters, which quantifies the global alignment between estimated and ground-truth camera trajectories.

All experiments are conducted on an Intel i9-13900K CPU with an NVIDIA RTX 4090 GPU. Vision modules execute in parallel on CUDA streams, while the SLAM back end runs on the CPU.

\subsection{Datasets}
\textbf{Dynamic scenes.} We use the \texttt{freiburg3} sequences \texttt{walking\_xyz}, \texttt{walking\_rpy}, and \texttt{walking\_halfsphere}, which contain significant non-rigid human motion. High-precision motion-capture ground truth makes these sequences well-suited for benchmarking robustness under dynamic conditions~\cite{sturm2012benchmark}.

\textbf{Static scenes.} To assess geometric consistency and long-term scale stability, we use nine rigid indoor sequences from the \texttt{freiburg1} subset (\texttt{360}, \texttt{desk}, \texttt{desk2}, \texttt{floor}, etc.)~\cite{sturm2012benchmark}, which capture standard conditions without dynamic interference.

\subsection{Baselines}
We compare against both classical and modern SLAM pipelines. RGB-D methods include ORB-SLAM2~\cite{mur2017orb}, DynaSLAM~\cite{bescos2018dynaslam}, DG-SLAM~\cite{xu2024dg}, and RoDynSLAM~\cite{jiang2024rodyn}. Monocular baselines span sparse and dense formulations, including DSO~\cite{engel2017direct}, DROID-SLAM~\cite{teed2021droid}, and recent approaches incorporating dynamic modeling or Gaussian splatting such as WildGS-SLAM~\cite{zheng2025wildgs} and DynaMon~\cite{schischka2024dynamon}.

\subsection{Performance in Static Environments}
\autoref{tab:tum_static_full} reports results on the static subset. Our approach achieves a mean ATE of 7.4\,cm with UniDepthV2~\cite{piccinelli2025unidepthv2} and 8.3\,cm with DepthAnythingV2~\cite{yang2024depth}. While competitive, these results are surpassed by fully learned, end-to-end pipelines such as MASt3R-SLAM~\cite{murai2025mast3r} (3.0\,cm) and DROID-SLAM~\cite{teed2021droid} (3.8\,cm), which jointly optimize depth, pose, and optical flow in a dense bundle adjustment framework. Such coupling enables correction of systematic errors in initial depth predictions, achieving higher global consistency.

Interestingly, even the strongest monocular methods outperform our RGB-D baseline (6.8\,cm ATE). This discrepancy stems from the structured-light depth sensor used in TUM~\cite{sturm2012benchmark}, which produces sparse and noisy depth on reflective or textureless surfaces. Learned monocular models, by contrast, infer dense, coherent geometry in these regions, effectively inpainting missing structure and yielding more stable trajectories.

Unlike end-to-end pipelines, our modular system does not refine depth estimates during optimization. While this limits correction of systematic errors, it highlights the strength of pretrained monocular depth networks, which already provide cues sufficient for competitive performance in a classical feature-based SLAM framework.

\subsection{Performance in Dynamic Environments}
\begin{table}[t]
  \centering
  \caption{ATE (m) on the TUM RGB-D~\cite{sturm2012benchmark} Dynamic benchmark. Lower is better. Values marked with (*) are from WildGS-SLAM~\cite{zheng2025wildgs}.}
  \label{tab:tum_dynamic_m}
  \begin{tabular}{lccc}
    \toprule
    Method & f3/w/xyz & f3/w/rpy & f3/w/hs \\
    \midrule
    \multicolumn{4}{@{}l}{\emph{RGB-D Baselines}}\\
    ReFusion~\cite{palazzolo2019refusion}   & 0.099 & 0.406\textsuperscript{*} & 0.104 \\
    ORB-SLAM2~\cite{mur2017orb}             & 0.722 & 0.805 & 0.723 \\
    DynaSLAM (N+G)~\cite{bescos2018dynaslam} & \cellcolor{yellow!20}0.015 & \cellcolor{green!20}0.035 & \cellcolor{yellow!20}0.025 \\
    NICE-SLAM~\cite{zhu2022nice}            & 0.865 & 2.440 & 1.520 \\
    DG-SLAM~\cite{xu2024dg}                 & 0.016 & 0.043 & -- \\
    RoDyn-SLAM~\cite{jiang2024rodyn}        & 0.083 & -- & 0.056 \\
    DDN-SLAM (RGB-D)~\cite{li2025ddn}       & \cellcolor{green!20}0.014 & \cellcolor{yellow!20}0.039 & \cellcolor{green!20}0.023 \\
    \rowcolor{gray!12}\textbf{Ours (RGB-D)} & \cellcolor{green!20}\textbf{0.012} & \cellcolor{green!20}\textbf{0.025} & \cellcolor{green!20}\textbf{0.018} \\
    \midrule
    \multicolumn{4}{@{}l}{\emph{Monocular Baselines}}\\
    DSO~\cite{engel2017direct}              & 0.129 & 0.138 & 0.407 \\
    DROID-SLAM~\cite{teed2021droid}         & 0.016 & 0.040 & 0.022 \\
    MonoGS~\cite{matsuki2024gaussian}       & 0.215 & 0.174 & 0.442 \\
    Splat-SLAM~\cite{sandstrom2025splat}    & \cellcolor{green!20}0.013 & 0.039 & 0.022 \\
    DynaMoN (MS)~\cite{schischka2024dynamon}     & \cellcolor{yellow!20}0.014 & 0.039 & \cellcolor{yellow!20}0.020 \\
    DynaMoN (MS\&SS)~\cite{schischka2024dynamon} & \cellcolor{green!20}0.014 & \cellcolor{green!20}0.031 & 0.019 \\
    DDN-SLAM (RGB)~\cite{li2025ddn}         & 0.028 & 0.089 & 0.041 \\
    MonST3R~\cite{zhang2024monst3r}         & 0.273 & 0.136 & 0.198 \\
    MegaSaM~\cite{li2025megasam}            & 0.015 & \cellcolor{green!20}\textbf{0.026} & \cellcolor{green!20}0.018 \\
    WildGS-SLAM~\cite{zheng2025wildgs}      & \cellcolor{green!20}0.013 & 0.033 & \cellcolor{green!20}\textbf{0.016} \\
    \rowcolor{gray!12}\textbf{Ours (w/DepthAnythingV2)} & \cellcolor{green!20}\textbf{0.013} & 0.034 & 0.025 \\
    \rowcolor{gray!12}\textbf{Ours (w/UniDepthV2)}      & \cellcolor{green!20}\textbf{0.013} & \cellcolor{yellow!20}0.032 & 0.023 \\
    \bottomrule
  \end{tabular}
\end{table}

Results on dynamic sequences are shown in \autoref{tab:tum_dynamic_m}. Our RGB-D configuration achieves 1.8\,cm mean ATE, outperforming all prior RGB-D methods, including specialized approaches such as DynaSLAM~\cite{bescos2018dynaslam} (2.8\,cm) and DDN-SLAM~\cite{li2025ddn} (2.5\,cm). Crucially, this improvement is achieved without explicit motion modeling or scene-flow estimation. Instead, robust instance segmentation with mask dilation ensures that only features from static regions are passed to the back end, yielding highly reliable correspondences.

The monocular variant using UniDepthV2~\cite{piccinelli2025unidepthv2} achieves 2.27\,cm mean ATE, surpassing all existing monocular pipelines and most RGB-D baselines. This demonstrates that combining reliable learned depth priors with aggressive semantic filtering can bridge the performance gap between monocular and depth-sensing systems.

The strength of our approach lies in its modular simplicity. By supplying a clean, metrically scaled set of static features to a standard geometric back end, we avoid the fragility of explicitly modeling dynamic motion. This design delivers state-of-the-art accuracy in dynamic environments while maintaining robustness across diverse scenes.

\subsection{Ablation Studies}
\label{sec:ablations}

We perform ablation experiments on the \texttt{f3/w/xyz} sequence to evaluate the contributions of our core modules. Unless otherwise noted, all experiments use the same hardware setup.

\subsubsection{Component Contribution and Runtime}
Removing instance masks increases ATE from 0.0135\,m to 0.270\,m (\autoref{tab:runtime_profile}), confirming that semantic filtering is indispensable in dynamic environments. Without filtering, the back end is dominated by incorrect correspondences from moving objects, leading to rapid tracking failure.  

\begin{figure}[t]
\centering
\begin{subfigure}[t]{0.49\linewidth}
\includegraphics[width=\linewidth]{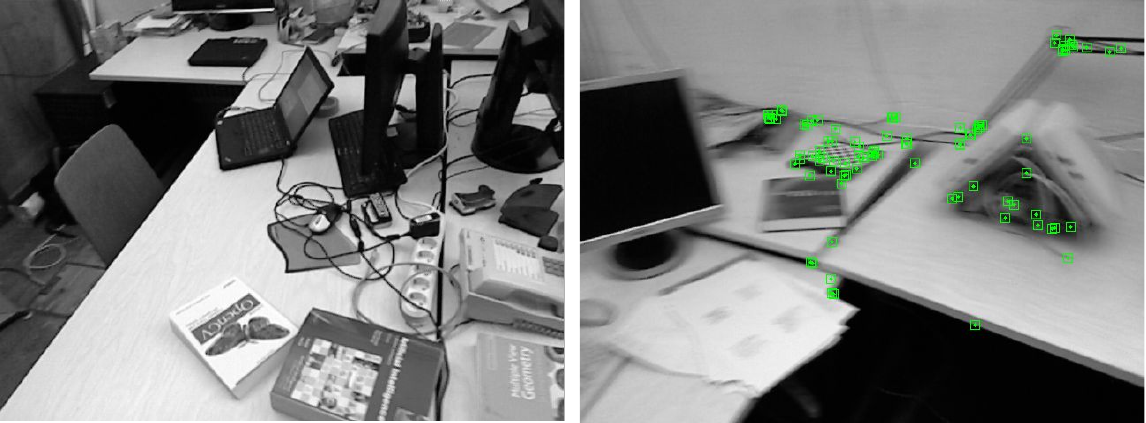}
\caption{ORB features~\cite{rublee2011orb}}
\end{subfigure}
\hfill
\begin{subfigure}[t]{0.49\linewidth}
\includegraphics[width=\linewidth]{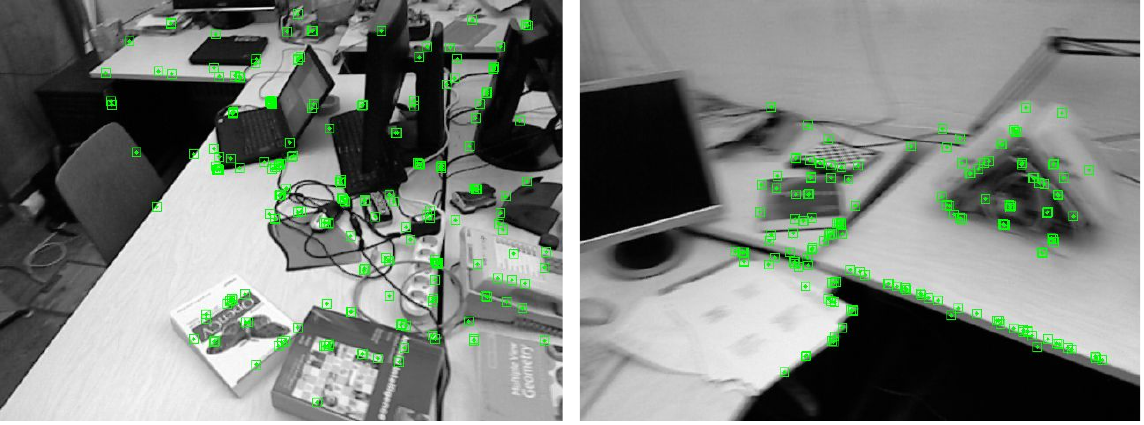}
\caption{Key.Net features~\cite{barroso2019key}}
\end{subfigure}
\caption{Feature detection comparison on a dynamic frame. ORB~\cite{rublee2011orb} features cluster in textured areas, whereas Key.Net~\cite{barroso2019key} produces a more uniform distribution, improving robustness under low texture and motion.}
\label{fig:combined-figure}
\end{figure}

Replacing Key.Net~\cite{barroso2019key} with ORB~\cite{rublee2011orb} increases ATE by 7\%. As shown in \autoref{fig:combined-figure}, ORB~\cite{rublee2011orb} features cluster in textured regions, whereas Key.Net~\cite{barroso2019key} provides more uniform coverage, improving stability in low-texture and motion-heavy conditions.  

Substituting monocular depth with ground-truth depth improves ATE only marginally (0.0135\,m $\rightarrow$ 0.0127\,m), showing that modern monocular estimators already provide sufficiently precise metric cues.  

\begin{table}[t]
  \centering
  \caption{Runtime ablation on \texttt{fr3\_walking\_xyz}. Each row disables or replaces one module of the full pipeline to measure its impact on accuracy (ATE in meters) and throughput (FPS). Higher FPS and lower ATE indicate better performance.}
  \label{tab:runtime_profile}
  \resizebox{\columnwidth}{!}{
  \begin{tabular}{lcc}
    \toprule
    Configuration & FPS & ATE (m) \\
    \midrule
    Full system & \textbf{22} & \textbf{0.0135} \\
    w/o YOLOv11~\cite{khanam2024yolov11} masks & 23 & 0.2700 \\
    w/o depth estimator (RGB-D)~\cite{yang2024depth,piccinelli2025unidepthv2} & \textbf{36} & 0.0127 \\
    ORB keypoints instead of Key.Net~\cite{barroso2019key} & 22 & 0.0145 \\
    \bottomrule
  \end{tabular}
  }
\end{table}

\begin{figure}[t]
\centering
\includegraphics[width=\columnwidth]{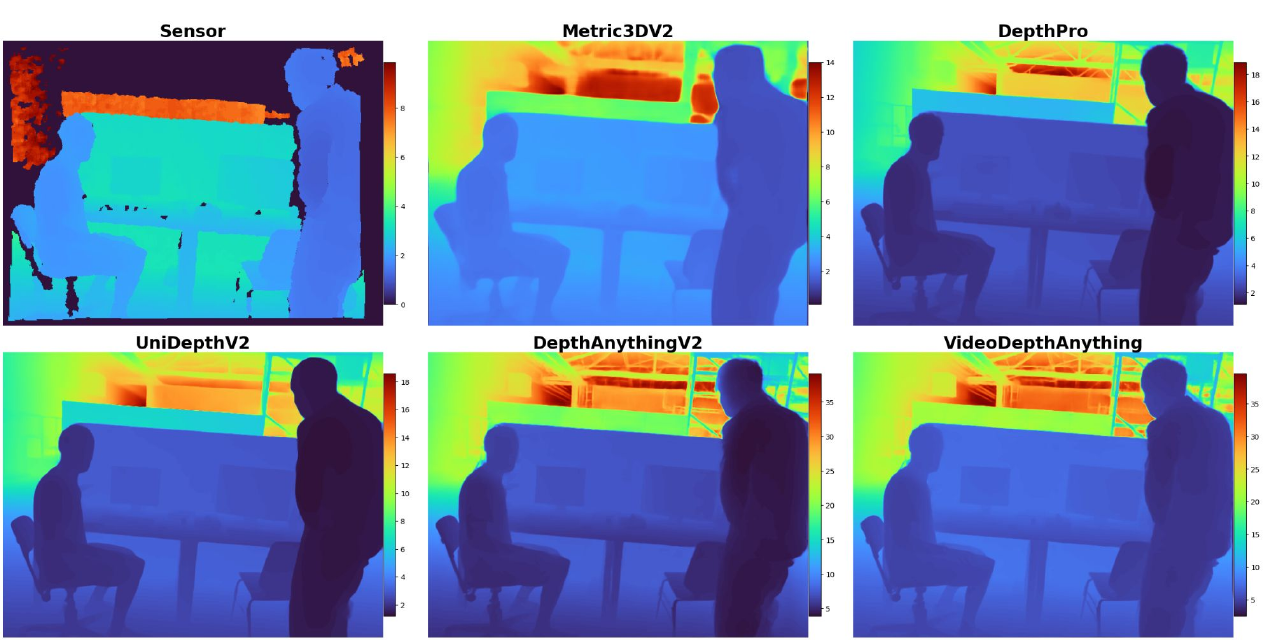}
\caption{Depth predictions from different models, individually rescaled for visualization. Differences highlight variations in relative structure and spatial consistency.}
\label{fig:depth_own_scale}
\end{figure}

\begin{table*}[t]
  \centering
  \caption{Comparison of monocular depth estimators under different scaling strategies on \texttt{fr3\_walking\_xyz}. Columns correspond to per-frame oracle scaling, range clipping at different thresholds, global calibration, and raw predictions. Values are ATE (m), where lower is better. ``X'' indicates failure cases.}
  \label{tab:depth_ablation_full}
  \resizebox{\textwidth}{!}{%
  \begin{tabular}{lcccccccc}
    \toprule
    \multirow{2}{*}{Depth estimator} &
      \multirow{2}{*}{Per-frame} &
      \multicolumn{5}{c}{Depth truncated to …} &
      \multirow{2}{*}{Global} &
      \multirow{2}{*}{Original} \\
    \cmidrule(lr){3-7}
      & & 3\,m & 5\,m & 7.5\,m & 10\,m & 15\,m & & \\
    \midrule
    VideoDepthAnything~\cite{chen2025video}     & \textbf{0.0148} & X & \textbf{0.0142} & 0.0145 & 0.0157 & 0.0158 & 0.0164 & 0.0160 \\
    UniDepthV2~\cite{piccinelli2025unidepthv2}  & 0.0186 & \textbf{0.0135} & 0.0146 & 0.0177 & 0.0188 & 0.0176 & 0.0174 & 0.0173 \\
    DepthPro~\cite{bochkovskii2024depth}        & 0.0168 & 0.0153 & 0.0159 & 0.0160 & 0.0161 & 0.0160 & 0.0162 & 0.0157 \\
    Metric3DV2~\cite{hu2024metric3d}            & 0.0239 & 0.0156 & 0.0161 & 0.0216 & 0.0230 & 0.0231 & 0.0219 & 0.0234 \\
    DepthAnythingV2~\cite{yang2024depth}        & 0.0195 & X & X & \textbf{0.0137} & \textbf{0.0142} & \textbf{0.0146} & \textbf{0.0140} & 0.0142 \\
    Depth sensor                                & --     & --     & --     & --     & --     & --     & --     & \textbf{0.0129} \\
    \bottomrule
  \end{tabular}}
\end{table*}

\begin{table}[t]
  \centering
  \caption{Effect of feature budget on SLAM accuracy (median ATE RMSE in meters) across eight TUM static sequences. Lower is better.}
  \label{tab:feature_budget}

  \resizebox{\columnwidth}{!}{
  \begin{tabular}{lcccccc}
    \toprule
    \multirow{2}{*}{\textbf{Sequence}} &
      \multicolumn{3}{c}{\textbf{UniDepthV2~\cite{piccinelli2025unidepthv2}}} &
      \multicolumn{3}{c}{\textbf{DepthAnythingV2~\cite{yang2024depth}}} \\
    \cmidrule(lr){2-4}\cmidrule(lr){5-7}
    \textbf{Features} & 1000 & 2000 & 3000 & 1000 & 2000 & 3000 \\
    \midrule
    floor   & 0.736 & 0.276 & 0.148 & 0.191 & 0.283 & \textbf{0.100} \\
    rpy     & 0.026 & 0.043 & 0.045 & 0.042 & \textbf{0.023} & \textbf{0.023} \\
    plant   & 0.036 & \textbf{0.035} & 0.043 & 0.079 & 0.071 & 0.075 \\
    360     & 0.160 & 0.115 & 0.117 & 0.102 & \textbf{0.094} & 0.099 \\
    teddy   & 0.059 & \textbf{0.055} & 0.092 & 0.198 & 0.227 & 0.310 \\
    room    & 0.069 & \textbf{0.058} & 0.081 & 0.080 & 0.074 & 0.099 \\
    desk2   & 0.039 & 0.036 & 0.038 & 0.029 & \textbf{0.026} & 0.027 \\
    desk    & 0.044 & 0.043 & 0.043 & 0.032 & \textbf{0.027} & \textbf{0.027} \\
    \midrule
    \textit{Average} & 0.145 & 0.088 & \textbf{0.076} & 0.094 & 0.102 & 0.095 \\
    \bottomrule
  \end{tabular}
  }
\end{table}

\begin{figure}[t]
\centering
\includegraphics[width=\columnwidth]{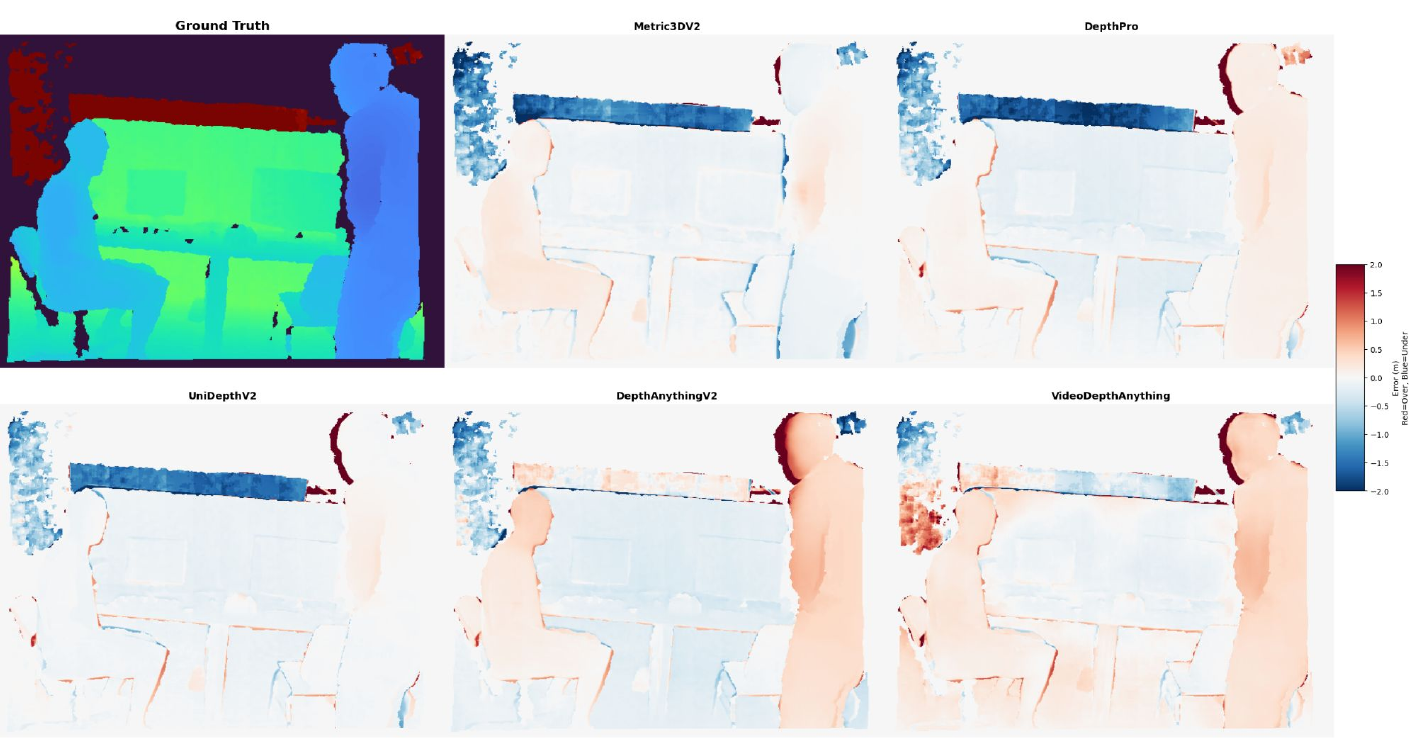}
\caption{Depth error maps for different models. Red regions indicate overestimation, blue indicates underestimation relative to ground truth.}
\label{fig:depth_error}
\end{figure}

\begin{table*}[t]
\centering
\caption{Temporal consistency analysis of monocular depth estimators on \texttt{fr3\_walking\_xyz}. Metrics include per-frame accuracy (RMSE), coefficient of variation (CV) for scale and error stability, and final SLAM error (ATE). Lower values indicate better performance.}
\label{tab:slam_performance_full}
  \resizebox{\textwidth}{!}{%
\begin{tabular}{lcccccc}
\toprule
Method & RMSE (m) & Scale CV & RMSE CV & MAE CV & AbsRel CV & ATE (m) \\
\midrule
DepthAnythingV2~\cite{yang2024depth} & 4.453 & \textbf{0.0493} & 0.367 & \textbf{0.367} & 0.591 & \textbf{0.0142} \\
DepthPro~\cite{bochkovskii2024depth} & 4.938 & 0.0524 & \textbf{0.344} & 0.406 & 0.594 & 0.0157 \\
VideoDepthAnything~\cite{chen2025video} & 4.865 & 0.0896 & 0.427 & 0.513 & 0.591 & 0.0160 \\
UniDepthV2~\cite{piccinelli2025unidepthv2} & 4.776 & 0.0508 & 0.376 & 0.408 & 0.662 & 0.0173 \\
Metric3DV2~\cite{hu2024metric3d} & \textbf{3.477} & 0.0648 & 0.385 & 0.372 & \textbf{0.495} & 0.0234 \\
\bottomrule
\end{tabular}
}
\end{table*}

The runtime profile in \autoref{tab:runtime_profile} shows that our pipeline sustains real-time performance above 20\,FPS. Increasing the feature budget from 1000 to 3000 has negligible impact on throughput, demonstrating efficient scaling with hardware.

\subsubsection{Depth Scaling Strategies and Error Structure}
Monocular depth predictors output relative depth up to an arbitrary scale. To integrate them into metric SLAM, we evaluated five models across four scaling strategies: (1) per-frame oracle scaling with ground-truth depth, (2) range clipping (3–15\,m), (3) global offline calibration, and (4) raw predictions.  

Range clipping consistently yielded the most stable performance. UniDepthV2~\cite{piccinelli2025unidepthv2} achieved optimal results at 3.0\,m clipping (0.0135\,m ATE), while DepthAnythingV2~\cite{yang2024depth} performed best at 7.5\,m (0.0137\,m ATE). Several models failed entirely at certain thresholds (marked ``X'' in \autoref{tab:depth_ablation_full}), underscoring the sensitivity of SLAM pipelines to depth configuration.  

Qualitative visualizations confirm systematic biases. \autoref{fig:depth_own_scale} shows that even metric-trained models predict distinct scale ranges and structures. Error maps in \autoref{fig:depth_error} reveal consistent tendencies: Metric3DV2~\cite{hu2024metric3d} overestimates human figures while underestimating backgrounds, whereas DepthAnythingV2~\cite{yang2024depth} and UniDepthV2~\cite{piccinelli2025unidepthv2} yield more balanced errors. Such structured biases, rather than raw accuracy, often determine downstream SLAM stability.

\subsubsection{Feature Budget Optimization}
The feature budget controls the number of keypoints retained per frame. Results in \autoref{tab:feature_budget} reveal model-specific effects.  

For UniDepthV2~\cite{piccinelli2025unidepthv2}, increasing the budget from 1000 to 3000 features reduces ATE from 0.145\,m to 0.076\,m, showing consistent gains from additional constraints. DepthAnythingV2~\cite{yang2024depth} exhibits non-monotonic behavior: the 2000-feature setting (0.102\,m) performs worse than both 1000 (0.094\,m) and 3000 (0.095\,m), suggesting overfitting or bottlenecks. These results confirm that optimal feature counts must be tuned per depth model.

\subsubsection{Temporal Consistency and SLAM Error}
Beyond per-frame accuracy, we assess the temporal stability of depth predictions, which is critical for sequential SLAM. Standard metrics such as RMSE, MAE, and AbsRel capture instantaneous error, while their coefficient of variation (CV) across frames quantifies stability:
\begin{equation}
\text{CV}(M) = \frac{\sigma(M)}{\mu(M)}.
\end{equation}
Low CV values indicate consistent predictions over time. In practice, \textit{Scale CV} is most relevant for preventing drift, \textit{RMSE CV} supports stable bundle adjustment, \textit{MAE CV} stabilizes residuals for outlier rejection, and \textit{AbsRel CV} preserves relative depth ratios.  

This reflects the sequential nature of SLAM: errors accumulate not only through their magnitude at each frame $\epsilon_t$ but also through temporal correlations. The variance of the cumulative error expands as
\begin{equation}
\mathrm{Var}\!\left(\sum_{t=1}^{N}\epsilon_t\right) 
= \sum_{t=1}^{N}\mathrm{Var}(\epsilon_t) + 2\sum_{i<j}\mathrm{Cov}(\epsilon_i,\epsilon_j),
\end{equation}
which shows that correlated errors amplify drift much more than independent ones. If errors are uncorrelated, the covariance terms vanish and uncertainty grows like a random walk, i.e., $\sqrt{N}$ in expectation. By contrast, systematic biases or temporally correlated residuals cause linear error growth, a phenomenon well documented in SLAM drift analyses~\cite{dubbelman2012bias, kerl2013robust, hidalgo2017gaussian}.  

As reported in \autoref{tab:slam_performance_full}, Metric3DV2~\cite{hu2024metric3d} achieves the lowest per-frame RMSE but exhibits high scale variation, leading to substantially worse SLAM error (0.0234\,m ATE). DepthAnythingV2~\cite{yang2024depth}, by contrast, achieves the best overall result (0.0142\,m ATE) through superior temporal stability despite higher instantaneous error. These results highlight our central conclusion: temporal consistency is more critical than per-frame accuracy for monocular SLAM.
\section{Discussion and Limitations}
\label{sec:discussion}

Our experiments show that pretrained vision models can raise monocular SLAM to the accuracy of RGB-D pipelines. By treating monocular depth prediction~\cite{yang2024depth,piccinelli2025unidepthv2}, instance segmentation~\cite{khanam2024yolov11}, and learned keypoints~\cite{barroso2019key} as plug-and-play virtual sensors, the system achieves metric scale and robustness to dynamic content in a zero-shot manner. Among these, instance-level masking proves most critical, as it prevents catastrophic tracking failures in the presence of human motion or other non-rigid dynamics. Depth priors eliminate scale drift and allow direct SE(3) pose estimation without additional calibration, while learned keypoints contribute to stability in texture-poor or motion-heavy scenes.

The modular design provides three practical benefits. First, it enables deployment without task-specific training or fine-tuning, making the system immediately applicable to unseen environments. Second, it maintains full compatibility with existing SLAM back ends, which lowers the barrier to integration into established robotics stacks. Third, it ensures longevity, since future improvements in pretrained vision models can be incorporated as drop-in replacements without retraining or redesigning the pipeline.

The approach also inherits limitations from its underlying modules. Depth and segmentation networks can degrade under domain shifts such as outdoor lighting, reflective or transparent materials, or previously unseen object categories. Depth is currently treated as a fixed observation without explicit uncertainty modeling, which allows systematic biases in scale or structure to propagate through the map. Scenarios dominated by moving objects or with insufficient static geometry remain especially challenging, as even robust filtering cannot guarantee stable tracking.

Our implementation is currently optimized for real-time sparse tracking on high-end GPUs and validated primarily on indoor datasets.

\section{Conclusion}
\label{sec:conclusion}

We present a monocular SLAM system that achieves metric-scale accuracy in static and dynamic indoor scenes using a single RGB camera by integrating pretrained modules for depth estimation, learned keypoints, and instance segmentation with an unmodified geometric back end. Our approach delivers results on the TUM RGB-D~\cite{sturm2012benchmark} benchmark that are competitive with RGB-D systems and state-of-the-art for dynamic monocular SLAM, while the modular architecture enables zero-shot deployment and maintains compatibility with existing SLAM back ends. This work represents an important shift in 3D perception, where learned depth priors increasingly rival dedicated hardware sensors, democratizing high-fidelity spatial awareness and enabling single-camera systems to attain the environmental understanding once limited to complex, costly multi-sensor configurations. Our results demonstrate that modern pretrained vision models can effectively replace active depth sensors for robust, metric-scale SLAM, opening new possibilities for cost-effective and power-efficient autonomous systems, and as the capabilities of pretrained vision models continue to advance, we anticipate that learned depth priors will become increasingly competitive with hardware sensors, potentially making RGB-D SLAM an unnecessary complexity for many applications.
\bibliographystyle{IEEEtran} 
\bibliography{IEEEabrv, reflist} 

\end{document}